\documentclass{article}
\usepackage{spconf,amsmath,amssymb,graphicx,hyperref,booktabs}
\usepackage{multirow}
\usepackage{float}

\title{DISENTANGLING LINGUISTIC AND PARALINGUISTIC INFORMATION WITH ROUTED SPARSE AUTOENCODERS}
\name{Beimnet Bekele Guta, Xiaoyu Yang, Guangzhi Sun, Philip C. Woodland}
\address{Department of Engineering, University of Cambridge, Cambridge, UK}
\begin{document}
%
\maketitle

\begin{abstract} Self-supervised speech encoders contain linguistic and paralinguistic information in a shared, entangled representation space. We combine a TopK sparse autoencoder with route-specific supervision and cross-factor adversaries. Across frozen SPEAR and WavLM encoders, independent probes show factor-specific retention and suppression: linguistic information remains stronger in the linguistic route, while paralinguistic factors, including speaker identity, emotion, and prosody, are retained in the paralinguistic route and substantially reduced in the linguistic route. The route organisation learned on LibriSpeech persists on MSP-Podcast without representation-side retraining. Feature-space route interventions further transfer the swapped factor while largely preserving the information carried by the unchanged route. These results show consistent route-selective separation across encoders, corpora, independent probes, and representation-level interventions. \end{abstract}


\section{Introduction} Self-supervised learning (SSL) has become central to speech representation learning, enabling speech encoders to learn from large amounts of unlabelled audio and produce representations that support tasks including phone recognition, automatic speech recognition, speaker identification, emotion recognition, and prosody-related prediction \cite{baevski2020wav2vec,hsu2021hubert,chen2022wavlm, yangLargeScaleEvaluationSpeech2024}. However, these different properties exist in a shared and entangled representation space. The same activation can carry phonetic content, speaker identity, prosody, and other information, making selective interpretation, removal, or modification of individual properties difficult. Hence, separating linguistic and paralinguistic information would support clearer interpretation and more selective control of speech representations.

Speech representation disentanglement has been addressed using perturbation-based invariance, information bottlenecks, architectural mechanisms, and adversarial training \cite{qianCONTENTVECImprovedSelfSupervised, qianUnsupervisedSpeechDecomposition2021, luSpeechTripleNetEndtoEndDisentangled2023, ganinDomainAdversarial2016,wangAdversariallyLearningDisentangled2021}. Sparse autoencoders (SAEs) provide a complementary approach by mapping a dense representation into a wider sparse code and reconstructing the original representation from a small subset of active units. Previous work has identified interpretable units in language, speech, and audio representations \cite{bricken2023monosemanticity,parraInterpretableSparseFeatures, mariotteSparseAutoencodersMake2025,aparinAudioSAETowards2026}. However, sparsity alone does not ensure that the learned units correspond to factors selected in advance. SAE dictionaries can depend on model width and training choices, and a reconstruction objective need not recover a canonical or predefined factorisation \cite{leaskSparseAutoencodersNot2025, pengUseSparseAutoencoders2025}. Existing speech and audio SAE studies primarily analyse discovered features rather than routing the sparse code into task-specific subspaces under an explicit disentanglement objective, which is the gap addressed here.

In this work, we investigate whether the sparse code generated by an SAE on top of an SSL speech encoder can be organised into linguistic and paralinguistic routes with factor-specific retention and suppression. The linguistic route retains phonetic and transcript information, while the paralinguistic route retains speaker identity, emotion, and prosody. Phone recognition (PR) provides the linguistic supervision signal, while speaker identification (SID), emotion recognition (ER), and prosody prediction provide the paralinguistic signals.
The proposed disentanglement framework combines a TopK sparse autoencoder \cite{makhzani2014ksparse,gaoScalingEvaluatingSparse2024} with unit-level routing, route-specific task supervision, and cross-factor adversarial objectives. 
The TopK SAE provides an overcomplete sparse representation, while unit-level routing, task supervision, and adversarial objectives organise its units into linguistic and paralinguistic routes.

We evaluate the framework using frozen SPEAR~\cite{yangSPEARUnifiedSSL2026} and WavLM~\cite{chen2022wavlm}representations \cite{panayotovLibriSpeechASR2015,yangSPEARUnifiedSSL2026, chen2022wavlm}. Across LibriSpeech and MSP-Podcast, independent probes evaluate linguistic information through PR and paralinguistic information through SID, ER, and prosody prediction. The evaluation includes reconstruction-only, fixed- routing, and quota-freeze SAEs, together with Eta-WavLM \cite{ruggiero-etal-2025-eta}, adversarial ablations, frozen cross-corpus transfer, and controlled route interventions.


\section{Methodology} \label{sec:method} We start with a frozen SSL speech encoder that acts as a source of the input representation. Its representation is fed into a TopK SAE, whose active units are partitioned into linguistic and paralinguistic routes. The SAE, routing masks, and task heads are trained jointly under reconstruction and task supervision losses, while the speech encoder is frozen.

\begin{figure}[t] 
\centering 
\includegraphics[width=\columnwidth]{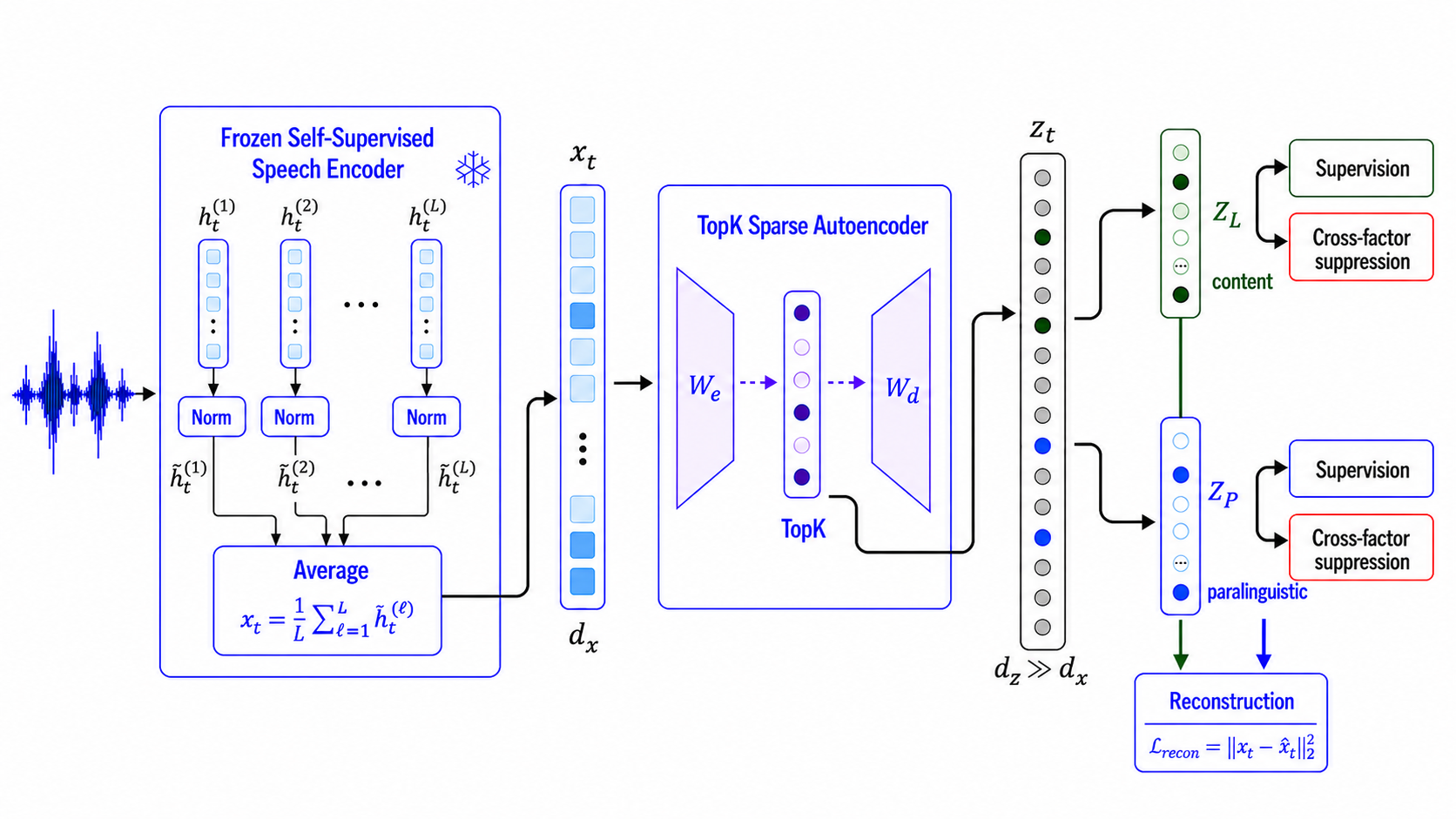} 
\vspace{-1.5em}
\caption{TopK SAE produce $\mathbf{z}_t$ on top of a frozen speech encoder, which is routed into linguistic $\mathbf{z}_L$ and paralinguistic $\mathbf{z}_P$ representations.} 
\label{fig:framework} 
\vspace{-1em}
\end{figure} 

\vspace{-0.3em}
\subsection{Sparse Autoencoder} \label{sec:sae} 
Given an input waveform, the frozen speech encoder produces frame-level representations $\mathbf{X} = \mathbf{x}_{1:T}$, where $T$ is the sequence length and $\mathbf{x}_t\in\mathbb{R}^{D}$. A linear SAE projects this representation into an overcomplete space while enforcing sparsity by selecting the TopK elements \cite{makhzani2014ksparse,gaoScalingEvaluatingSparse2024}. The encoder and decoder are defined as \begin{align} \mathbf{s}_t &= \mathbf{W}_{\mathrm{enc}}\mathbf{x}_t +\mathbf{b}_{\mathrm{enc}}, \label{eq:sae_pre}\\ \mathbf{z}_t &= \operatorname{TopK}(\mathbf{s}_t;k), \label{eq:sae_topk}\\ \hat{\mathbf{x}}_t &= \mathbf{W}_{\mathrm{dec}}\mathbf{z}_t +\mathbf{b}_{\mathrm{dec}}, \label{eq:sae_dec} \end{align} where $\mathbf{W}_{\mathrm{enc}}\in\mathbb{R}^{K\times D}$, $\mathbf{W}_{\mathrm{dec}}\in\mathbb{R}^{D\times K}$, $K>D$, and $k$ is the active-unit budget. The reconstruction objective is \begin{equation} \mathcal{L}_{\mathrm{recon}} = \mathbb{E}_t \left\| \mathbf{x}_t-\hat{\mathbf{x}}_t \right\|_2^2. \label{eq:recon} \end{equation} TopK provides exact control over the number of active units without introducing shrinkage on the retained activations. To prevent dead activations, AuxK uses inactive units to predict the reconstruction residual \cite{gaoScalingEvaluatingSparse2024}: \begin{equation} \mathcal{L}_{\mathrm{AuxK}} = \mathbb{E}_t \left\| (\mathbf{x}_t-\hat{\mathbf{x}}_t) - \mathbf{W}^{(\mathrm{aux})}_{\mathrm{dec}} \mathbf{z}^{(\mathrm{aux})}_t \right\|_2^2. \label{eq:auxk} \end{equation} Here, $\mathbf{W}^{(\mathrm{aux})}_{\mathrm{dec}}$ and $\mathbf{z}^{(\mathrm{aux})}_t$ are restricted to the selected inactive units. This encourages them to acquire useful structure and re-enter the main TopK selection. 

\vspace{-0.5em}
\subsection{Task-Specific Routing} \label{sec:routing} 
Without supervision, SAE units might not align with the factors of interest \cite{leaskSparseAutoencodersNot2025,pengUseSparseAutoencoders2025}. Each latent unit is therefore mapped to either the linguistic route or the paralinguistic route with learnable masks $\mathbf{m}_L$ and $\mathbf{m}_P$ over the latent space: 
\begin{align} 
\mathbf{z}_L &= \mathbf{m}_L\odot\mathbf{z}_t, & \mathbf{z}_P &= \mathbf{m}_P\odot\mathbf{z}_t, \label{eq:routed_z}\\ m_L[u]+m_P[u] &=1, && u\in\{1,\ldots,K\}. \label{eq:mask_constraint} 
\end{align} 

The linguistic route $\mathbf{z}_L$ is trained to retain linguistic information while reducing paralinguistic factor recoverability and the paralinguistic route $\mathbf{z}_P$ is trained to retain paralinguistic information while reducing linguistic recoverability. For learned routing, the masks are global vectors shared across frames and utterances. Each unit $u$ has trainable routing logits $\mathbf{a}_u$, and its assignment is obtained using straight-through hard Gumbel--Softmax \cite{jangCategoricalReparameterization2017}: \begin{equation} [m_L[u],m_P[u]] = \operatorname{STGumbelSoftmax}(\mathbf{a}_u;\tau). \label{eq:gumbel} \end{equation} Writing $p_{u,\ell}=\operatorname{softmax}(\mathbf{a}_u)_\ell$, the routing loss encourages a decisive assignment: \begin{equation} \mathcal{L}_{\mathrm{routing}} = -\frac{1}{K} \sum_{u=1}^{K} \sum_{\ell\in\{L,P\}} p_{u,\ell}\log p_{u,\ell}. \label{eq:routing_loss} \end{equation} 
\textbf{Fixed routing} predetermines unit membership and applies separate budgets $k_L$ and $k_P$, where $k_L+k_P=k$. Both membership and budgets remain fixed throughout training. \textbf{Quota-freeze routing} first learns global assignments while a global TopK selects $k$ activations per frame. At the freeze step, the assignments are fixed, and their mean activity is converted into separate route quotas $k_L$ and $k_P$. Training then continues with separate TopK selection within each route. 

\subsection{Task Supervision and Adversarial Training} \label{sec:supervision} 
Task heads attached to each route create supervision signals that align it with its intended factors. Phone recognition supervises $\mathbf{z}_L$, while speaker identification, emotion classification, and prosody prediction supervise $\mathbf{z}_p$.

\vspace{-0.4em}
\begin{align}
\mathcal{L}_{\mathrm{PR}} &= \operatorname{CTC} \left( h_{\mathrm{PR}}(\mathbf{z}_L),y_{\mathrm{ph}} \right), \label{eq:pr_loss}\\ \mathcal{L}_{\mathrm{SID}} &= -\log p_{\phi} \left( y_{\mathrm{spk}} \mid \operatorname{pool}(\mathbf{z}_P) \right), \label{eq:sid_loss}\\ \mathcal{L}_{\mathrm{Pros}} &= \operatorname{ProsMSE} \left( h_{\mathrm{Pros}}(\mathbf{z}_P),y_{\mathrm{Pros}} \right), \label{eq:pros_loss}\\ \mathcal{L}_{\mathrm{Emo}} &= -w_e\log p_{\phi} \left( y_{\mathrm{Emo}} \mid \operatorname{pool}(\mathbf{z}_P) \right). \label{eq:emo_loss} 
\end{align}
Here, $y_{\mathrm{ph}}$ denotes the transcript-derived phone target sequence, $y_{\mathrm{spk}}$ and $y_{\mathrm{Emo}}$ denote the utterance-level speaker and emotion labels, and $y_{\mathrm{Pros}}$ contains frame-level log-$F_0$ and mean-centred log-energy. The operator $\operatorname{pool}(\cdot)$ denotes temporal mean pooling, and $w_e$ is the emotion-class weight.

These supervision signals alone do not guarantee disentanglement. A phone adversary therefore acts on $\mathbf{z}_P$, while speaker, emotion, and prosody adversaries act on $\mathbf{z}_L$. The gradient reversal layer acts as the identity during the forward pass and multiplies the backward gradient by $-\gamma$ \cite{ganinDomainAdversarial2016}: \begin{equation} \operatorname{GRL}_{\gamma}(\mathbf{z})=\mathbf{z}, \qquad \frac{\partial\operatorname{GRL}_{\gamma}} {\partial\mathbf{z}} = -\gamma\mathbf{I}. \label{eq:grl} \end{equation}

For brevity, let
$\mathbf{a}_{\mathrm{PR}}
=\operatorname{GRL}_{\gamma_{\mathrm{PR}}}(\mathbf{z}_P)$
and
$\mathbf{a}_t
=\operatorname{GRL}_{\gamma_t}(\mathbf{z}_L)$
for $t\in\{\mathrm{SID,Pros,Emo}\}$.
The adversarial losses are:
\vspace{-0.2em}
\begin{align}
\mathcal{L}^{(P)}_{\mathrm{adv,PR}}
&= \operatorname{CTC}
(h_{\mathrm{adv,PR}}(\mathbf{a}_{\mathrm{PR}}),
y_{\mathrm{ph}}),
\label{eq:adv_pr}\\
\mathcal{L}^{(L)}_{\mathrm{adv,SID}}
&= -\log p_{\psi}
(y_{\mathrm{spk}}\mid
\operatorname{pool}(\mathbf{a}_{\mathrm{SID}})),
\label{eq:adv_sid}\\
\mathcal{L}^{(L)}_{\mathrm{adv,Pros}}
&= \operatorname{ProsMSE}
(h_{\mathrm{adv,Pros}}(\mathbf{a}_{\mathrm{Pros}}),
y_{\mathrm{Pros}}),
\label{eq:adv_pros}\\
\mathcal{L}^{(L)}_{\mathrm{adv,Emo}}
&= -w_e\log p_{\psi}
(y_{\mathrm{Emo}}\mid \notag\\
&\quad h_{\mathrm{adv,Emo}}
(\operatorname{pool}(\mathbf{a}_{\mathrm{Emo}}))).
\label{eq:adv_emo}
\end{align}
The total training objective is
\begin{equation} \mathcal{L}_{\mathrm{train}} = \lambda_{\mathrm{recon}}\mathcal{L}_{\mathrm{recon}} + \lambda_{\mathrm{AuxK}}\mathcal{L}_{\mathrm{AuxK}} + \mathcal{L}_{\mathrm{task}} + \mathcal{L}_{\mathrm{adv}}, \label{eq:total_loss} \end{equation} 
with $\mathcal{L}_{\mathrm{routing}}$ added during learned routing. Since the masks are complementary, $\mathbf{z}_t=\mathbf{z}_L+\mathbf{z}_P$, so the complete sparse code continues to reconstruct the encoder representation.

\vspace{-0.4em}
\section{Experimental Setup} \label{sec:setup} 
\vspace{-0.2em}
\subsection{Data, Encoders, and Configurations} 

The experiments use LibriSpeech \texttt{train-clean-100} \cite{panayotovLibriSpeechASR2015} and MSP-Podcast \cite{bussoMSPPodcastCorpus2026}. Phone labels are obtained from transcripts using the CMU pronunciation lexicon with a grapheme-to-phoneme fallback, producing a 74-symbol phone vocabulary. On LibriSpeech, PR uses \texttt{dev-clean} for validation and \texttt{test-clean} for evaluation. Closed-set SID uses utterance-disjoint splits within \texttt{train-clean-100}, with the same 251 speakers across splits. MSP-Podcast evaluates linguistic information through PR, and paralinguistic information through SID, emotion, and prosody.

We evaluate frozen SPEAR XLarge \cite{yangSPEARUnifiedSSL2026} and WavLM Large \cite{chen2022wavlm}. For both encoders, parameter-free LayerNorm is applied per layer before uniform averaging. The upstream encoders remain frozen throughout SAE training, and all runs use the same architecture, routing, optimisation settings. 
All SAEs use dictionary width $K=5120$, TopK $k=256$, AuxK budget 64, and AuxK weight $1/32$.  Fixed routing uses contiguous $K_L/K_P=4096/1024$ assigned units and active budgets $k_L/k_P=240/16$, selected in preliminary experiments from the validation trade-off between intended-factor retention and cross-factor suppression.

Under quota-freeze routing, assignments are learned until step 4,000 and the mean active split over 20 sampled training batches sets the post-freeze budgets. The SPEAR reconstruction-only and fixed-routing models are trained for 12,000 steps. The corresponding WavLM configurations use 12,000 steps, except quota-freeze with 16,000 steps. We additionally evaluate Eta-WavLM, which fits an affine speaker component from ECAPA-TDNN embeddings and subtracts it from the WavLM representation \cite{ruggiero-etal-2025-eta}.

\vspace{-0.5em}
\subsection{Evaluation Protocol} 
All probes are trained post hoc on frozen representations, with parameters entirely separate from the SAE, routing masks, and training-time adversaries. Relying on the training-time adversary as evidence that information has been removed is unreliable \cite{elazarAdversarialRemoval2018}. PR is evaluated using phone error rate (PER), SID using accuracy. Emotion is evaluated using unweighted average recall (UAR), while prosody is evaluated using $F_0$ and energy correlations. Reconstruction fidelity is measured by the mean-squared error between $\mathbf{x}_t$ and $\hat{\mathbf{x}}_t$. For cross-corpus evaluation, the frozen LibriSpeech-trained models are evaluated on a 51,017-utterance MSP-Podcast subset with 700 speakers and 41,067/4,975/4,975 train/validation/test utterances; closed-set SID chance is $0.143\%$ \cite{bussoMSPPodcastCorpus2026}. The SAE encoder and decoder, routing masks, and quotas remain frozen; only MSP-specific phone and speaker probes are trained. 

For controlled intervention, we use a fixed manifest of 250 held-out recipient--donor utterance pairs with different speakers and at most 10\% duration difference, reused across both routing models. The donor route is linearly interpolated over normalised utterance time to match the recipient's number of frames. A $P$-route swap combines the recipient's $\mathbf{z}_L$ with the resampled donor $\mathbf{z}_P$, while an $L$-route swap combines the resampled donor $\mathbf{z}_L$ with the recipient's $\mathbf{z}_P$. The recombined code is passed through the frozen SAE decoder, producing a reconstructed SPEAR feature $\hat{\mathbf{h}}$ with the recipient's frame count. Independent phone and speaker evaluators, fitted only on disjoint unswapped reconstructions, operate on $\hat{\mathbf{h}}$. 
We report the percentage of samples retaining the recipient phone after a $P$-swap, changing from the recipient phone after an $L$-swap, or specifically matching the donor phone after an $L$-swap. Speaker-match scores report the percentage of samples predicted as the donor speaker after a $P$-swap or retained as the recipient speaker after an $L$-swap.

\begin{table*}[!t]
    \centering
    \caption{Independent probes across encoders and corpora.
    Reconstruction MSE (Recon.) is reported once per SAE model.}
    \vspace{0.2cm}
    \label{tab:main_results}
    \vspace{1pt}
    \setlength{\tabcolsep}{3.2pt}
    \renewcommand{\arraystretch}{1.0}
    \scriptsize
    \resizebox{\textwidth}{!}{%
    \begin{tabular}{lllrrrrrrrrr}
        \toprule
        & & &
        \multicolumn{3}{c}{LibriSpeech} &
        \multicolumn{6}{c}{MSP-Podcast} \\
        \cmidrule(lr){4-6}
        \cmidrule(lr){7-12}
        Encoder & SAE Model & Route
        & Recon.\,$\downarrow$
        & PER\,$\downarrow$
        & SID Acc.\,$\uparrow$
        & Recon.\,$\downarrow$
        & PER\,$\downarrow$
        & SID Acc.\,$\uparrow$
        & Emo. UAR\,$\uparrow$
        & $F_0$ corr.\,$\uparrow$
        & Energy corr.\,$\uparrow$ \\
        \midrule

        \multirow{7}{*}{SPEAR X-Large}
        & Recon-only & $\mathbf{z}_t$
        & 0.066 & 6.00 & 99.6
        & 0.075 & 32.37 & 61.74 & 65.25 & 0.578 & 0.944 \\
        \cmidrule(lr){2-12}
        & \multirow{3}{*}{Fixed} & $\mathbf{z}_t$
        & 0.179 & 10.50 & 100.0
        & 0.146 & 30.52 & 65.98 & 66.25 & 0.482 & 0.923 \\
        & & $\mathbf{z}_L$
        & & 10.02 & 6.2
        & & 30.58 & 20.67 & 51.71 & 0.118 & 0.781 \\
        & & $\mathbf{z}_P$
        & & 99.60 & 99.8
        & & 81.79 & 76.47 & 67.40 & 0.564 & 0.918 \\
        \cmidrule(lr){2-12}
        & \multirow{3}{*}{Quota-freeze} & $\mathbf{z}_t$
        & 0.163 & 8.73 & 100.0
        & 0.171 & 34.74 & 68.16 & 67.46 & 0.485 & 0.924 \\
        & & $\mathbf{z}_L$
        & & 8.94 & 0.6
        & & 35.01 & 32.29 & 60.26 & 0.236 & 0.833 \\
        & & $\mathbf{z}_P$
        & & 80.26 & 99.8
        & & 89.15 & 73.83 & 67.92 & 0.575 & 0.919 \\
        \midrule

        \multirow{7}{*}{WavLM Large}
        & Recon-only & $\mathbf{z}_t$
        & 0.043 & 6.84 & 100.0
        & 0.048 & 30.86 & 60.79 & 58.94 & 0.592 & 0.962 \\
        \cmidrule(lr){2-12}
        & \multirow{3}{*}{Fixed} & $\mathbf{z}_t$
        & 0.113 & 10.58 & 99.6
        & 0.097 & 30.63 & 57.07 & 57.92 & 0.443 & 0.957 \\
        & & $\mathbf{z}_L$
        & & 10.68 & 0.8
        & & 31.45 & 18.62 & 47.95 & 0.077 & 0.803 \\
        & & $\mathbf{z}_P$
        & & 94.81 & 99.6
        & & 79.00 & 69.38 & 60.12 & 0.524 & 0.956 \\
        \cmidrule(lr){2-12}
        & \multirow{3}{*}{Quota-freeze} & $\mathbf{z}_t$
        & 0.108 & 9.19 & 100.0
        & 0.104 & 31.81 & 66.83 & 60.06 & 0.432 & 0.959 \\
        & & $\mathbf{z}_L$
        & & 9.40 & 14.8
        & & 31.90 & 29.19 & 53.56 & 0.211 & 0.895 \\
        & & $\mathbf{z}_P$
        & & 80.84 & 99.8
        & & 80.05 & 70.55 & 60.67 & 0.517 & 0.956 \\
        \midrule

        \multirow{2}{*}{Eta-WavLM}
        & \multirow{2}{*}{Raw WavLM} & $\mathbf{h}$
        & -- & 6.08 & 97.35
        & -- & 28.82 & 74.86 & 59.73 & 0.637 & 0.978 \\
        & & $\mathbf{h}_{\eta}$
        & -- & 7.25 & 95.41
        & -- & 29.80 & 66.22 & 55.11 & 0.589 & 0.977 \\
        \bottomrule
    \end{tabular}%
    }
    \vspace{-2em}
\end{table*}

\vspace{-0.5em}
\section{Results} 
\vspace{-0.4em}
\label{sec:results} 
\subsection{Linguistic and Paralinguistic Separation} \label{sec:separation} 

Table~\ref{tab:main_results} shows factor-specific retention and suppression across encoders, corpora, and tasks. All SAE models are trained either on LibriSPeech or MSP-Podcast. The reconstruction-only SAEs preserve linguistic and paralinguistic information, but sparsity alone does not produce the selected factorisation. Under both SPEAR and WavLM, phone information remains stronger in $\mathbf{z}_L$, while speaker identity remains stronger in $\mathbf{z}_P$. 
The MSP-Podcast results show the same route organisation across the broader task set. Phone information remains stronger in $\mathbf{z}_L$, while speaker identity, emotion, and prosody are more recoverable from $\mathbf{z}_P$. Across encoders, fixed routing provides stronger paralinguistic suppression in $\mathbf{z}_L$, whereas quota-freeze more strongly suppresses linguistic information in $\mathbf{z}_P$ and provides a more balanced reconstruction and route-selectivity trade-off. In the separate Eta-WavLM evaluation, the affine residual largely preserves phone performance but provides weaker factor-specific suppression than the routed representations.

Beyond the models trained directly on each corpus, we additionally test whether the LibriSpeech-trained SPEAR routing can transfer to MSP-Podcast without fine-tuning. Fixed $\mathbf{z}_L/\mathbf{z}_P$ yields 43.92/99.72\% PER and 12.82/82.71\% SID accuracy, while quota-freeze gives 44.56/90.38\% PER and 3.00/82.29\% SID accuracy, both retaining the opposing route pattern under corpus shift.

\vspace{-0.5em}
\subsection{Controlled Route Interventions} 
\vspace{-0.2em}
\label{sec:interventions} 

Table~\ref{tab:swaps} shows a similar double dissociation under both routing configurations. Swapping $\mathbf{z}_P$ preserves recipient phone information while transferring donor speaker identity, whereas swapping $\mathbf{z}_L$ transfers donor phone information while preserving recipient speaker identity. The swap therefore provides strong evidence that the routes are functionally distinct rather than merely statistically distinct.

\begin{table}[t] 
\centering 
\caption{Controlled route interventions on the fixed 250-pair SPEAR evaluation set (\%).} \label{tab:swaps} 
\vspace{0.2cm}
\setlength{\tabcolsep}{3.2pt} \renewcommand{\arraystretch}{0.9} 
\begin{tabular}{lrr} 
\toprule
Outcome & Fixed & Freeze \\
\midrule
$P$-swap phone retention        & 99.57 & 99.16 \\
$P$-swap donor-speaker match    & 98.00 & 97.20 \\
$L$-swap phone replacement      & 94.26 & 94.12 \\
$L$-swap recipient-speaker match & 96.00 & 96.80 \\
$L$-swap donor-phone retention  & 99.10 & 98.84 \\
\bottomrule
\end{tabular} 
\vspace{-0.7em}
\end{table} 

\subsection{Ablations} 
\label{sec:ablations}

Table~\ref{tab:ablations} shows that task supervision retains the desired information. Without adversaries, $\mathbf{z}_L$ retains strong phone performance and $\mathbf{z}_P$ retains near-ceiling speaker accuracy. However, cross-factor information also remains recoverable. The speaker adversary reduces speaker recoverability from $\mathbf{z}_L$, while the phone adversary reduces phonetic recoverability from $\mathbf{z}_P$. The two provide complementary suppression.

\begin{table}[t]
\centering
\caption{Ablations of the adversarial objectives on SPEAR. Lower $\mathbf{z}_L$ SID and higher $\mathbf{z}_P$ PER indicate stronger cross-factor suppression.}
\vspace{0.2cm}
\label{tab:ablations}
\resizebox{\columnwidth}{!}{%
\begin{tabular}{llrrrr}
\toprule
\multirow{2}{*}{Routing} & \multirow{2}{*}{Condition} & $\mathbf{z}_L$ PER & $\mathbf{z}_P$ PER & $\mathbf{z}_L$ SID & $\mathbf{z}_P$ SID \\
 & & (\%) & (\%) & (\%) & (\%) \\
\midrule
Fixed  & Full model      & 10.02 & 99.60 & 6.2   & 98.6  \\
       & No adversaries  & 4.49  & 66.64 & 99.8  & 100.0 \\
       & No speaker adv. & 4.63  & 85.25 & 100.0 & 99.4  \\
       & No phone adv.   & 11.01 & 47.82 & 1.6   & 100.0 \\
\midrule
Freeze & Full model      & 8.94  & 80.26 & 0.6   & 99.6  \\
       & No adversaries  & 4.75  & 12.36 & 99.8  & 100.0 \\
       & No speaker adv. & 5.00  & 86.10 & 99.8  & 99.2  \\
       & No phone adv.   & 9.90  & 15.04 & 0.6   & 99.6  \\
\bottomrule
\end{tabular}%
}
\vspace{-0.8em}
\end{table}

\section{Conclusion} 
This paper investigated the disentanglement of pretrained speech representations using SAE. With suitable inductive biases, an SAE can produce strong route-selective separation. Route-specific task supervision, cross-factor adversaries, and a stable routing boundary separate the SAE latent units into linguistic and paralinguistic routes with factor-specific retention and suppression. Independent probes show that each route retains the intended information, while cross-factor recoverability is substantially reduced. This pattern is observed with both SPEAR XLarge and WavLM Large and is retained under cross-corpus evaluation on MSP-Podcast. Route swapping further shows that the learned routes have distinct functional roles.


\vfill\pagebreak
\clearpage

\bibliographystyle{IEEEbib}
\bibliography{refs}

\end{document}